\documentclass[10pt,twocolumn,letterpaper]{article}

\usepackage[pagenumbers]{cvpr} %

\usepackage{microtype}

\renewcommand{\paragraph}[1]{\vspace{.5em}\noindent\textbf{#1.}}

\usepackage{amsmath, amsfonts, bm}

\usepackage{amsmath,amsfonts,bm}

\def\eqref#1{equation~\ref{#1}}

\def\1{\bm{1}}

\def\vb{{\bm{b}}}

\def\vp{{\bm{p}}}

\def\vu{{\bm{u}}}

\def\vw{{\bm{w}}}
\def\vx{{\bm{x}}}

\def\mS{{\bm{S}}}

\def\mW{{\bm{W}}}
\def\mX{{\bm{X}}}

\DeclareMathAlphabet{\mathsfit}{\encodingdefault}{\sfdefault}{m}{sl}
\SetMathAlphabet{\mathsfit}{bold}{\encodingdefault}{\sfdefault}{bx}{n}

\newcommand{\R}{\mathbb{R}}

\def\xp{\vx_{\vp}}

\definecolor{cvprblue}{rgb}{0.21,0.49,0.74}
\usepackage[pagebackref,breaklinks,colorlinks,allcolors=cvprblue]{hyperref}
\usepackage{caption}
\usepackage{subcaption}

\def\paperID{*****} %
\def\confName{CVPR}
\def\confYear{2026}

\title{Why Fake ? Unveiling the Semantic Vocabulary of Deepfake Detectors}

\author{
Vazgken Vanian\thanks{Equal contribution} \\
{\tt\small vvanian@iti.gr}
\and
Alexandros Doumanoglou\footnotemark[1] \\
{\tt\small aldoum@iti.gr}
\and
Dimitris Zarpalas \\
{\tt\small zarpalas@iti.gr} 
\and
Information Technologies Institute (ITI)\\
Centre For Research and Technology HELLAS (CERTH)
}

\begin{document}
\maketitle
\begin{abstract}
Deepfake (DF) technology poses a significant threat to information integrity, driving the need for robust detection methods. Most DF detectors only consider predicting a binary label for whether the input is real or fake, lacking the justification required for real-world applications like legal proceedings. Explainable DF Detection has emerged to address this limitation, but existing techniques frequently fall short by either relying on human annotations for precise artifact localization or generating superficially plausible textual explanations without grounding. This work investigates the use of post-hoc explainable AI (XAI) to analyze the decision-making process of state-of-the-art black-box DF detectors. Specifically, we employ Encoding-Decoding Direction Pairs (EDDP), a technique suitable for uncovering the concept space of DF detectors (their semantic vocabulary) as well as the mechanism for writing and reading concept information to and from internal representations. Our analysis reveals previously hidden \textit{real} and \textit{fake} features learned implicitly during detector training, offering nuanced explanations unattainable through conventional methods. This enables global model understanding, spatially aware concept localization, and counterfactual what-if analysis, all contributing to a deeper comprehension of DF detection strategies. 
\end{abstract}
    
\section{Introduction}
\label{sec:introduction}

The proliferation of deepfake (DF) content presents an escalating challenge to information integrity across various domains, from journalism and politics to legal proceedings \cite{khan2025comprehensive,chesney2019deep}.
As generative models continue to advance, realism has surpassed previously unimaginable thresholds, necessitating the development of robust detection methodologies. The majority of existing approaches focus on binary classification \cite{afchar2018mesonet,roessler2019faceforensicspp}. 
While effective, this approach suffers from a critical limitation: a lack of transparency and justification. In contexts demanding accountability a simple \textit{real} or \textit{fake} label is insufficient. Instead, understanding \textit{why} a piece of content is flagged as manipulated is paramount.

This need for explainability has spurred recent research into Explainable Deepfake Detection (XDFD). Existing approaches broadly fall into two categories: spatio-temporal localization  \cite{miao2023multi,chen2022self,hu2024delocate} and textual explanation methods \cite{hondru2025exddv}.

Spatio-temporal localization methods aim to identify manipulated regions by highlighting areas of the image or segments of the video that are presumed to contain forged content. The motivation is to move beyond binary prediction and provide visual evidence supporting the decision. However, in practice, these methods often produce coarse localization, frequently defaulting to highlighting the entire face rather than isolating specific manipulation artifacts. As a result, they provide limited insight into what precise cues led to the prediction.

Textual explanation methods instead generate natural language justifications describing why an image was classified as fake. While these explanations improve human interpretability, they often lack accurate spatial grounding and may not faithfully reflect the underlying evidence used by the detector. Consequently, despite their explanatory intent, existing XDFD methods primarily function as prediction tools with attached justification mechanisms, rather than systems that deliver precise and causally grounded explanations.

In this work, we explore XDFD from a different perspective by analyzing deepfake detectors through the lens of \textbf{post-hoc} Explainable Artificial Intelligence (XAI). Rather than modifying detectors to explicitly produce explanations, we aim to uncover and interpret the intrinsic features learned during standard training.

This perspective provides several advantages. First, explanations are derived directly from the representations that drive the detector’s prediction, ensuring tight alignment between explanation and decision function and thereby improving faithfulness. Second, because our approach is purely post-hoc, it introduces no architectural modifications or auxiliary training objectives, avoiding potential trade-offs between interpretability and predictive performance. Third, the framework can be applied to existing pretrained detectors without retraining or artifact-level supervision, making it deployable across models and datasets.

Our technical approach centers on Encoding-Decoding Direction Pairs (EDDP) \cite{doumanoglou2025learning}, a recent technique that can unveil a) the concepts that a deep vision network uses to make predictions (its semantic vocabulary), and b) the network’s mechanism for encoding and decoding these concepts in internal representations, under the linear representation hypothesis \cite{bereska2024mechanistic}. We argue that EDDP concepts are a natural fit for XDFD because access to the network’s encoding-decoding mechanism for real and fake concepts enables: (a) global model understanding, by identifying which concepts drive predictions toward \textit{real} or \textit{fake}; (b) spatially aware, concept-based local explanations; and (c) counterfactual what-if analysis.

To the best of our knowledge, this is the first application of post-hoc, concept-based XAI within the deepfake detection domain, opening a new avenue for improving transparency and trust in automated content verification systems.

\section{Related Work}
\label{sec:related-work}

\subsection{DeepFake Generation}

Deepfake synthesis methods can be broadly categorized into face-swapping, face-reenactment, and audio-driven lip-syncing.

\textbf{Face-swapping} replaces identity while preserving pose and expression. Early works relied on encoder–decoder frameworks and required person-specific training \cite{korshunova2017fast, liu2023deepfacelab}. Later many-to-many models such as SimSwap \cite{chen2020simswap} enabled identity transfer across unseen subjects. Subsequent works improved temporal coherence and realism using transformers and motion–appearance decoupling strategies \cite{kim2022smooth, cui2023face, luo2025canonswap}. More recently, diffusion-based approaches have emerged \cite{kim2025diffface, zhao2023diffswap, ye2025dreamid, wang2025dynamicface}. They leverage pretrained generative priors for high-fidelity synthesis.

\textbf{Face-reenactment} transfers motion and expression from a driving source to a target identity. Face2Face \cite{thies2016face2face} pioneered 3D model-based reenactment, followed by learning-based approaches that emphasize generalization and one-shot transfer \cite{bounareli2023hyperreenact, rochow2024fsrt}. Large-scale diffusion video models further improve realism and temporal stability \cite{qiu2025skyreels}.

\textbf{Lip-syncing} methods generate mouth motion from speech signals. Wav2Lip \cite{prajwal2020lip} introduced adversarial synchronization learning, while SadTalker \cite{zhang2023sadtalker} models 3D motion coefficients for natural head dynamics. Diffusion-based systems such as \cite{guan2024resyncer} extend this paradigm toward holistic audio-driven human video generation.

\subsection{Deepfake Detection and Datasets}

Deepfake detection has evolved alongside advances in generation techniques. Early methods relied on hand-crafted cues, such as abnormal eye blinking or head-pose inconsistencies \cite{li2018ictu, yang2019exposing}. CNN-based approaches like MesoNet \cite{afchar2018mesonet} and XceptionNet \cite{roessler2019faceforensicspp} captured subtle spatial artifacts, while frequency-aware methods such as F3-Net \cite{qian2020thinking} exploited frequency-domain inconsistencies. Transformer-based architectures later integrated spatial and temporal frequency representations for improved robustness \cite{kim2025beyond}, and recent CLIP-based approaches leverage vision-language representations with parameter-efficient fine-tuning to enhance cross-dataset generalization \cite{yan2024effort, yermakov2025unlocking}.

Datasets have grown in parallel to match these advances, from early benchmarks like UADFV \cite{yang2019exposing} and Deepfake-TIMIT to FaceForensics++ (FF++) \cite{roessler2019faceforensicspp} and Celeb-DF \cite{li2020celeb}, which improved realism and temporal consistency. Larger-scale collections such as DFDC \cite{dolhansky2020deepfake}, DeeperForensics-1.0 \cite{jiang2020deeperforensics}, ForgeryNet \cite{he2021forgerynet}, and recent benchmarks DF40 \cite{yan2024df40} and DDL \cite{miao2025ddl} provide greater diversity and coverage of modern generation techniques. These datasets form the foundation for evaluating robust and generalizable deepfake detectors.

\subsection{Explainability in DeepFake Detection}

Explainability for deepfake detection remains comparatively underexplored, with most existing works relying on post-hoc saliency maps like Grad-CAM \cite{selvaraju2017grad} that provide limited insight into model reasoning. Recent research has introduced several distinct strategies to address this. ProtoExplorer \cite{bouter2024protoexplorer} utilizes prototype-based reasoning within a human-in-the-loop framework, allowing predictions to be grounded in learned visual archetypes. In a different approach, the Locally-Explainable Self-Blended (LESB) detector \cite{soltandoost2025extracting} focuses on Local Feature Discovery to decompose global predictions into discrete, region-level contributions. This is achieved with part specific self-blending augmentations that focus on key areas like eyes,nose and mouth.

Multimodal transparency is explored in ExDDV \cite{hondru2025exddv}, which pairs spatial annotations with textual justifications tailored for vision-language models. Furthermore, network dissection is utilized in \cite{mansoor2025explainable} to quantify transparency by aligning individual internal CNN neurons with semantic facial concepts. 

While these individual methods advance the interpretability of detection models, they often introduce significant operational overhead. For instance, the network dissection approach in \cite{mansoor2025explainable} requires fine-tuning standard architectures, such as VGG-16, Inception V3, or ResNet-50, on forensic datasets to map internal activations to meaningful concepts. Similarly, methods like ProtoExplorer \cite{bouter2024protoexplorer} impose specific architectural constraints through the requirement of specialized prototype layers. These dependencies suggest that high interpretability currently relies on additional supervision or task-specific retraining, leaving open the challenge of developing truly scalable, model-intrinsic explanation mechanisms that can be applied to diverse, unlabelled datasets without architectural modification.

\section{Unveiling DeepFake Concepts with Encoding-Decoding Direction Pairs (EDDP)}
\label{sec:background}

To understand how deepfake detectors represent and utilize information about real and fake content, we leverage a technique called \textbf{Encoding-Decoding Direction Pairs (EDDP)}. A core assumption underlying EDDP is the \textbf{linear representation hypothesis} \cite{bereska2024mechanistic}. This hypothesis posits that concepts, in this case, features indicative of real or fake images, are encoded within image representations as linear combinations of concept embeddings, i.e., vectors in the latent space, each associated with a concept.

\subsection{Method}

Given a feature representation $\mX \in \R^{D \times H \times W}$ and its constituent patch embeddings $\xp \in \R^D$, EDDP jointly learns the matrix of decoding directions $\mW \in \R^{D \times I}$, the encoding directions $\mS \in \R^{D \times I}$, and a vector of thresholds $\vb \in \R^I$ for a predefined number of concepts $I \in \mathbb{N}^+$. 

The learned decoding directions $\mW$, coupled with thresholds $\vb$, serve as deepfake related concept detectors. For any patch $\xp$, the presence of concept $i$ is determined by the condition:
\begin{equation}
    \vw_i^T \xp - b_i > 0
\end{equation}
The encoding directions $\mS$ represent the concept embeddings, which denote the specific directions in the latent space along which each concept is represented.

\subsection{Concept Detection Maps and Concept Contribution Maps}

We localize each concept within an image using concept detectors. To do this, we generate a \textbf{Concept Detection Map} or \textbf{Concepts Presence Map (CPMs)}: first, the detector is applied to each spatial element of the image representation $\mX$, then a hard thresholding step creates a binary map, and finally, the map is up-sampled to match the original image resolution. While Concept Detection Maps identify the presence of a concept, they do not quantify that concept's actual influence on the model's final prediction. 

To quantify this influence, we compute \textbf{Concept Contribution Maps (CCMs)}. CCMs are based on a sample's concept contributions and baseline concept contributions to the explanation logit, which is the difference between the network's class logit for an input image and a class logit corresponding to a baseline artificial representation from the uncertainty region, where concept detectors are at their decision threshold, as defined by \cite{doumanoglou2025learning}. CCMs provide a spatially aware breakdown and quantify the contributions of each concept across patches.

Visualization examples for CPMs and CCMs are shown in Figure \ref{fig:cpm-ccm}.

\section{Concept Analysis and Global Understanding}
\label{sec:concept-analysis}

\begin{figure*}
    \centering
    \includegraphics[width=\linewidth]{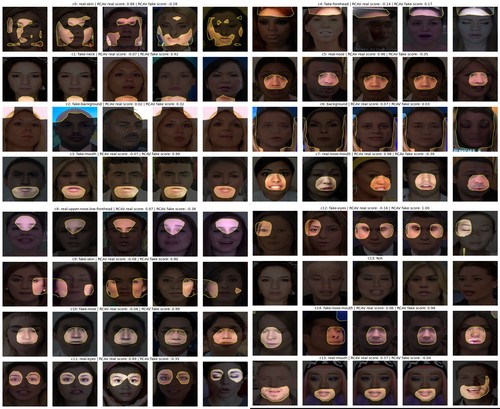}
    \caption{Visualization of the discovered concept vocabulary. Each concept $c_i$ is illustrated by representative face patches with high activation. The accompanying RCAV scores indicate the relative contribution of each concept toward a "real" or "fake" classification}
    \label{fig:concept_voc}
\end{figure*}

\subsection{Experimental Setup}

For our experimental framework, we employ an Xception \cite{chollet2017xception} architecture pre-trained on the FaceForensics++ (FF++) \cite{roessler2019faceforensicspp} dataset.  We apply the EDDP method to the activations of the model's 12th residual block, using the FF++ training set as our concept extraction source. 

The selection of the 12th residual block is motivated by a trade-off between semantic depth and the dimensionality of the feature space. While deeper layers represent higher-level semantic information, they often undergo a "rank collapse" as the network converges toward a single classification decision. Our empirical analysis using Principal Component Analysis (PCA) indicates that the 12th block maintains a sufficiently high-rank feature space compared to subsequent layers. PCA reveals that this layer maintains a high-rank feature space (rank $\approx 25$), capturing over 90\% of the variance. We empirically set the number of concepts to 16 concepts after evaluating the trade-off between concept-transfer accuracy, misclassified samples correction success, and semantic simplicity across settings of 12 to 24 concepts.

\subsection{Concept Identification}

The objective of our concept identification process is to isolate human-interpretable concepts that govern the model's final decision-making process. To characterize the learned concepts, establish their semantic groundedness, and analyze their distribution across the dataset, we employ three validation procedures: Relative Concept Activation Vector (RCAV) sensitivity analysis, semantic mapping, and distribution analysis.

It is important to note that the concepts identified through this process are intrinsically coupled with both the architectural priors of the Xception backbone and the specific data distribution of FF++. Because EDDP decomposes the model's internal latent space, the resulting encoding and decoding directions represent the specific features and artifacts that the detector has learned to prioritize for this particular classification task. Consequently, these concepts reflect the intersection of the model's representational capacity and the visual cues present within the FF++ dataset

We evaluate the functional influence of each identified concept on the model's decision-making process using RCAV \cite{pfau2021robust} sensitivity scores. As detailed in Table~\ref{tab:rcav_sensitivity}, these scores quantify the model's sensitivity to specific concepts $c_i$ with respect to the "Real" and "Fake" class predictions. Our analysis reveals that a subset of concepts ($c_1, c_3, c_9, c_{10}, c_{12}, c_{14}$) consistently biases the model's output toward the "Fake" class, serving as primary indicators of manipulation. Conversely, concepts such as $c_0, c_5, c_7, c_8,$ and $c_{11}$ strongly correlate with the "Real" class prediction. Notably, certain concepts ($c_2, c_4, c_6, c_{15}$) exhibit negligible impact on the final logit, suggesting they represent features that, while present, are not utilized as discriminative evidence by the detector.

Following this, we quantitatively map the concepts to specific facial regions using Interesection over Union (IoU) scores. These are calculated between concept presence maps and a ground-truth proxy derived from facial segmentation predictions. As detailed in Table \ref{tab:concept_class_iou}, this mapping allows us to associate abstract concept directions with interpretable semantic classes (e.g. $c_5$ showing high IoU with the "nose" region). 

Furthermore, we use the IoU metric, the RCAV sensitivity scores as well as qualitative inspection of the top-5 most strongly activated samples (as shown in Figure \ref{fig:concept_voc}) and assign descriptive names to the identified concepts. A summary of this naming is provided in Table \ref{tab:conceptmapping}.

\subsection{Concept Distribution and Manipulation Statistics}

To understand how these concepts manifest across the dataset, we calculate the dataset-wide concept presence as shown in Table \ref{tab:concept_presence}. This reveals the prevalence of specific features in real versus manipulated imagery. Further, we analyze the overlap between identified concepts and specific manipulation methods present in the FF++ dataset (FS,NT,DF,FF) in Table \ref{tab:concept_manipulation_overlap}.

Notably, there are certain concepts that exhibit high specificity to certain deepfake types. For instance, $c_3$ and $c_9$ show significantly higher presence in DeepFakes (DF) and Face2Face (FF) compared to real images, suggesting these concepts capture artifacts inherent to those specific generative processes. This distribution remains consistent across both training and testing splits, indicating the robustness of the identified concepts.

\begin{table}[ht]
\caption{RCAV sensitivity scores per concept on the test set. Higher positive scores indicate a stronger positive correlation with the respective class}
\label{tab:rcav_sensitivity}
\resizebox{\columnwidth}{!}{%
\begin{tabular}{lrrrrrrrr}
\toprule
 & \textbf{c0} & \textbf{c1} & \textbf{c2} & \textbf{c3} & \textbf{c4} & \textbf{c5} & \textbf{c6} & \textbf{c7} \\
\midrule
\textbf{Real} & 0.73 & -0.39 & -0.32 & -0.38 & -0.43 & 0.94 & -0.13 & 0.99 \\
\textbf{Fake} & -0.29 & 0.88 & 0.36 & 0.97 & 0.19 & -0.37 & 0.03 & -0.43 \\
\midrule
 & \textbf{c8} & \textbf{c9} & \textbf{c10} & \textbf{c11} & \textbf{c12} & \textbf{c13} & \textbf{c14} & \textbf{c15} \\
\midrule
\textbf{Real} & 0.98 & -0.35 & -0.32 & 0.84 & -0.48 & 0.98 & -0.20 & 0.09 \\
\textbf{Fake} & -0.44 & 0.88 & 0.97 & -0.35 & 1.00 & 0.79 & 0.96 & -0.05 \\
\bottomrule
\end{tabular}
}

\end{table}

\begin{table*}[ht]
\caption{Semantic alignment of learned concepts via IoU. We report the IoU between concept activation maps and ground-truth facial semantic masks. The highest overlap for each concept is highlighted in \textcolor{blue}{blue}, while non-zero overlaps are in \textcolor{red}{red}}
\label{tab:concept_class_iou}
\resizebox{\textwidth}{!}{%
\begin{tabular}{lccccccccccccccccccc}
\toprule
 & \textbf{background} & \textbf{skin} & \textbf{nose} & \textbf{eye\_g} & \textbf{l\_eye} & \textbf{r\_eye} & \textbf{l\_brow} & \textbf{r\_brow} & \textbf{l\_ear} & \textbf{r\_ear} & \textbf{mouth} & \textbf{u\_lip} & \textbf{l\_lip} & \textbf{hair} & \textbf{hat} & \textbf{ear\_r} & \textbf{neck\_l} & \textbf{neck} & \textbf{cloth} \\
\midrule
\textbf{c0} & \textcolor{red}{0.03} & \textbf{\textcolor{blue}{0.21}} & \textcolor{red}{0.01} & 0.00 & 0.00 & 0.00 & \textcolor{red}{0.01} & \textcolor{red}{0.01} & 0.00 & 0.00 & 0.00 & \textcolor{red}{0.01} & \textcolor{red}{0.01} & \textcolor{red}{0.05} & 0.00 & \textcolor{red}{0.01} & 0.00 & \textcolor{red}{0.04} & \textcolor{red}{0.03} \\
\textbf{c1} & 0.00 & 0.00 & 0.00 & 0.00 & 0.00 & 0.00 & 0.00 & 0.00 & 0.00 & 0.00 & 0.00 & 0.00 & 0.00 & 0.00 & 0.00 & 0.00 & 0.00 & \textbf{\textcolor{blue}{0.22}} & \textcolor{red}{0.05} \\
\textbf{c2} & \textbf{\textcolor{blue}{0.09}} & 0.00 & 0.00 & 0.00 & 0.00 & 0.00 & 0.00 & 0.00 & 0.00 & 0.00 & 0.00 & 0.00 & 0.00 & \textcolor{red}{0.05} & \textcolor{red}{0.01} & 0.00 & 0.00 & 0.00 & 0.00 \\
\textbf{c3} & 0.00 & \textcolor{red}{0.03} & 0.00 & 0.00 & 0.00 & 0.00 & 0.00 & 0.00 & 0.00 & 0.00 & \textcolor{red}{0.01} & \textcolor{red}{0.02} & \textbf{\textcolor{blue}{0.07}} & 0.00 & 0.00 & 0.00 & 0.00 & \textcolor{red}{0.01} & 0.00 \\
\textbf{c4} & \textcolor{red}{0.04} & \textcolor{red}{0.04} & 0.00 & 0.00 & 0.00 & 0.00 & 0.00 & 0.00 & 0.00 & 0.00 & 0.00 & 0.00 & 0.00 & \textbf{\textcolor{blue}{0.30}} & \textcolor{red}{0.01} & 0.00 & 0.00 & 0.00 & 0.00 \\
\textbf{c5} & 0.00 & \textcolor{red}{0.02} & \textbf{\textcolor{blue}{0.19}} & \textcolor{red}{0.01} & \textcolor{red}{0.01} & \textcolor{red}{0.01} & 0.00 & 0.00 & 0.00 & 0.00 & 0.00 & 0.00 & 0.00 & 0.00 & 0.00 & 0.00 & 0.00 & 0.00 & 0.00 \\
\textbf{c6} & \textbf{\textcolor{blue}{0.29}} & 0.00 & 0.00 & 0.00 & 0.00 & 0.00 & 0.00 & 0.00 & \textcolor{red}{0.02} & \textcolor{red}{0.02} & 0.00 & 0.00 & 0.00 & \textcolor{red}{0.18} & 0.00 & 0.00 & 0.00 & \textcolor{red}{0.04} & \textcolor{red}{0.16} \\
\textbf{c7} & 0.00 & \textcolor{red}{0.03} & \textbf{\textcolor{blue}{0.13}} & 0.00 & 0.00 & 0.00 & 0.00 & 0.00 & 0.00 & 0.00 & \textcolor{red}{0.01} & \textcolor{red}{0.05} & \textcolor{red}{0.02} & 0.00 & 0.00 & 0.00 & 0.00 & 0.00 & 0.00 \\
\textbf{c8} & 0.00 & \textbf{\textcolor{blue}{0.04}} & \textcolor{red}{0.01} & 0.00 & 0.00 & 0.00 & \textcolor{red}{0.02} & \textcolor{red}{0.02} & 0.00 & 0.00 & 0.00 & 0.00 & 0.00 & 0.00 & 0.00 & 0.00 & 0.00 & 0.00 & 0.00 \\
\textbf{c9} & \textcolor{red}{0.01} & \textbf{\textcolor{blue}{0.04}} & 0.00 & 0.00 & 0.00 & 0.00 & \textcolor{red}{0.01} & \textcolor{red}{0.01} & \textcolor{red}{0.01} & \textcolor{red}{0.01} & 0.00 & 0.00 & 0.00 & \textcolor{red}{0.01} & 0.00 & 0.00 & 0.00 & 0.00 & 0.00 \\
\textbf{c10} & 0.00 & \textcolor{red}{0.03} & \textbf{\textcolor{blue}{0.15}} & \textcolor{red}{0.01} & 0.00 & 0.00 & 0.00 & 0.00 & 0.00 & 0.00 & 0.00 & \textcolor{red}{0.02} & 0.00 & 0.00 & 0.00 & 0.00 & 0.00 & 0.00 & 0.00 \\
\textbf{c11} & 0.00 & \textcolor{red}{0.03} & 0.00 & \textcolor{red}{0.01} & \textcolor{red}{0.01} & \textcolor{red}{0.02} & \textcolor{red}{0.04} & \textbf{\textcolor{blue}{0.07}} & 0.00 & 0.00 & 0.00 & 0.00 & 0.00 & 0.00 & 0.00 & 0.00 & 0.00 & 0.00 & 0.00 \\
\textbf{c12} & 0.00 & \textbf{\textcolor{blue}{0.03}} & 0.00 & \textcolor{red}{0.01} & \textcolor{red}{0.01} & \textcolor{red}{0.01} & \textcolor{red}{0.03} & \textcolor{red}{0.03} & 0.00 & 0.00 & 0.00 & 0.00 & 0.00 & \textcolor{red}{0.01} & 0.00 & 0.00 & 0.00 & 0.00 & 0.00 \\
\textbf{c13} & 0.00 & 0.00 & 0.00 & 0.00 & 0.00 & 0.00 & 0.00 & 0.00 & 0.00 & 0.00 & 0.00 & 0.00 & 0.00 & 0.00 & 0.00 & 0.00 & 0.00 & 0.00 & 0.00 \\
\textbf{c14} & \textcolor{red}{0.03} & \textcolor{red}{0.02} & \textbf{\textcolor{blue}{0.09}} & 0.00 & 0.00 & 0.00 & 0.00 & 0.00 & 0.00 & 0.00 & \textcolor{red}{0.01} & \textcolor{red}{0.02} & \textcolor{red}{0.01} & \textcolor{red}{0.01} & 0.00 & 0.00 & 0.00 & 0.00 & 0.00 \\
\textbf{c15} & 0.00 & \textcolor{red}{0.03} & 0.00 & 0.00 & 0.00 & 0.00 & 0.00 & 0.00 & 0.00 & 0.00 & \textcolor{red}{0.02} & \textcolor{red}{0.05} & \textbf{\textcolor{blue}{0.09}} & 0.00 & 0.00 & 0.00 & 0.00 & 0.00 & 0.00 \\
\bottomrule
\end{tabular}

}

\end{table*}

\begin{table}[ht]
\caption{Concept ID to semantic label mapping. Labels are derived from a combination of the semantic IoU analysis \ref{tab:concept_class_iou} and qualitative inspection of high-activation patches}
\label{tab:conceptmapping}
\centering
\resizebox{\columnwidth}{!}{%
\begin{tabular}{llll}
\toprule
\textbf{c0}: real-skin & \textbf{c1}: fake-neck & \textbf{c2}: fake-background & \textbf{c3}: fake-mouth \\
\midrule
\textbf{c4}: fake-forehead & \textbf{c5}: real-nose & \textbf{c6}: background & \textbf{c7}: real-nose-mouth \\
\midrule
\textbf{c8}: real-upper-nose-low-forehead & \textbf{c9}: fake-skin & \textbf{c10}: fake-nose & \textbf{c11}: real-eyes \\
\midrule
\textbf{c12}: fake-eyes & \textbf{c13}: N/A & \textbf{c14}: fake-nose-mouth & \textbf{c15}: real-mouth \\
\bottomrule
\end{tabular}
}

\end{table}

\begin{table}[ht]
\caption{Global concept presence across the dataset. We report the percentage of images in which each concept $c_i$ is active}
\label{tab:concept_presence}
\centering
\small
\begin{tabular}{lcccc}
\toprule
 & \multicolumn{2}{c}{Train} & \multicolumn{2}{c}{Test} \\
\midrule
 & \textbf{Real (\%)} & \textbf{Fake (\%)} & \textbf{Real (\%)} & \textbf{Fake (\%)} \\
\midrule
\textbf{c0} & 21.8 & 78.2 & 21.6 & 78.4 \\
\textbf{c1} & 20.8 & 75.3 & 20.9 & 77.0 \\
\textbf{c2} & 10.7 & 38.3 & 11.2 & 38.6 \\
\textbf{c3} & 2.7 & 45.3 & 2.8 & 42.8 \\
\textbf{c4} & 21.7 & 78.1 & 21.6 & 78.4 \\
\textbf{c5} & 20.9 & 18.9 & 19.9 & 19.2 \\
\textbf{c6} & 21.8 & 78.2 & 21.6 & 78.4 \\
\textbf{c7} & 20.7 & 17.0 & 19.4 & 16.9 \\
\textbf{c8} & 19.5 & 30.1 & 18.2 & 31.5 \\
\textbf{c9} & 0.6 & 34.8 & 0.7 & 32.5 \\
\textbf{c10} & 0.1 & 35.0 & 0.2 & 33.3 \\
\textbf{c11} & 19.9 & 22.6 & 18.2 & 22.2 \\
\textbf{c12} & 1.7 & 43.3 & 1.9 & 40.4 \\
\textbf{c13} & 0.0 & 0.0 & 0.0 & 0.0 \\
\textbf{c14} & 4.1 & 34.5 & 5.1 & 35.2 \\
\textbf{c15} & 20.3 & 20.3 & 19.3 & 22.7 \\
\bottomrule
\end{tabular}

\end{table}

\begin{table}[ht]
\caption{Concept overlap by manipulation type and real samples. We break down concept activations across four specific forgery techniques: FaceSwap (FS), NeuralTextures (NT), DeepFake (DF), and Face2Face (FF).}
\label{tab:concept_manipulation_overlap}
\centering
\small
\begin{tabular}{lccccc}
\toprule
\multicolumn{6}{c}{Train} \\
\midrule
 & \textbf{Real (\%)} & \textbf{FS (\%)} & \textbf{NT (\%)} & \textbf{DF (\%)} & \textbf{FF (\%)} \\
\midrule
\textbf{c0} & 100.0 & 100.0 & 100.0 & 100.0 & 100.0 \\
\textbf{c1} & 95.7 & 95.6 & 97.3 & 96.0 & 96.4 \\
\textbf{c2} & 49.1 & 48.7 & 49.4 & 49.5 & 48.5 \\
\textbf{c3} & 12.4 & 30.7 & 17.4 & 90.3 & 79.3 \\
\textbf{c4} & 99.8 & 99.7 & 99.9 & 99.7 & 99.9 \\
\textbf{c5} & 96.2 & 10.0 & 91.5 & 3.8 & 2.1 \\
\textbf{c6} & 100.0 & 100.0 & 100.0 & 100.0 & 100.0 \\
\textbf{c7} & 95.3 & 9.4 & 84.6 & 1.4 & 1.6 \\
\textbf{c8} & 89.5 & 16.7 & 85.3 & 8.6 & 48.5 \\
\textbf{c9} & 2.7 & 4.3 & 3.9 & 74.5 & 79.0 \\
\textbf{c10} & 0.3 & 70.6 & 1.6 & 95.1 & 8.1 \\
\textbf{c11} & 91.6 & 6.8 & 88.3 & 3.5 & 24.5 \\
\textbf{c12} & 8.0 & 72.0 & 10.3 & 75.4 & 58.0 \\
\textbf{c13} & 0.0 & 0.0 & 0.0 & 0.0 & 0.0 \\
\textbf{c14} & 19.0 & 30.5 & 19.9 & 32.6 & 85.5 \\
\textbf{c15} & 93.4 & 28.3 & 71.0 & 4.1 & 10.0 \\
\midrule
\multicolumn{6}{c}{Test} \\
\midrule
 & \textbf{Real (\%)} & \textbf{FS (\%)} & \textbf{NT (\%)} & \textbf{DF (\%)} & \textbf{FF (\%)} \\
\midrule
\textbf{c0} & 100.0 & 100.0 & 100.0 & 100.0 & 100.0 \\
\textbf{c1} & 96.8 & 97.6 & 98.8 & 98.9 & 97.9 \\
\textbf{c2} & 51.6 & 49.1 & 51.7 & 47.3 & 49.4 \\
\textbf{c3} & 13.0 & 31.5 & 14.9 & 81.7 & 78.7 \\
\textbf{c4} & 99.9 & 100.0 & 100.0 & 100.0 & 99.9 \\
\textbf{c5} & 91.8 & 14.3 & 83.5 & 5.4 & 3.9 \\
\textbf{c6} & 100.0 & 100.0 & 100.0 & 100.0 & 100.0 \\
\textbf{c7} & 89.5 & 13.2 & 76.4 & 2.8 & 2.8 \\
\textbf{c8} & 84.3 & 27.3 & 78.2 & 13.8 & 46.2 \\
\textbf{c9} & 3.4 & 5.9 & 4.0 & 66.4 & 75.6 \\
\textbf{c10} & 1.1 & 60.5 & 3.1 & 91.4 & 10.9 \\
\textbf{c11} & 84.3 & 12.5 & 80.2 & 6.5 & 21.0 \\
\textbf{c12} & 8.7 & 64.5 & 11.3 & 70.1 & 55.2 \\
\textbf{c13} & 0.0 & 0.0 & 0.0 & 0.0 & 0.0 \\
\textbf{c14} & 23.6 & 35.9 & 24.0 & 32.4 & 81.7 \\
\textbf{c15} & 89.4 & 33.1 & 71.6 & 6.7 & 13.1 \\
\bottomrule
\end{tabular}

\end{table}

\section{Explanation Faithfulness Assessment}
\label{sec:faithfulness}

To evaluate the faithfulness of the learned concept explanations, we introduce a concept cloning intervention that directly manipulates the classifier's internal representation and measures whether the resulting predictions change in a controlled and predictable manner. If the learned concepts are genuinely integrated into the model’s decision process, then intervening on the concept coefficients should systematically affect the model output.

We begin by decomposing the intermediate representation at a selected layer. Specifically, we express the representation as a base component $\mX_{bc}$, which lies in the uncertainty region of all concepts, and a coefficient vector $\vu$ that quantifies the deviation from this concept-neutral point along each learned signal direction.

To assess faithfulness, we sample pairs of inputs $(\mX_s, \mX_t)$, referred to as source and target examples, and compute their respective decompositions:
\begin{equation}
\mX_s = \mX_{bc,s} + \mS^\top \vu_s,
\label{eq:source_decomp}
\end{equation}
\begin{equation}
\mX_t = \mX_{bc,t} + \mS^\top \vu_t.
\label{eq:target_decomp}
\end{equation}

We then construct a synthetic representation by combining the source base component with the target concept coefficients:
\begin{equation}
\mX_{\mathrm{syn}} = \mX_{bc,s} + \mS^\top \vu_t.
\label{eq:synthetic_rep}
\end{equation}

This operation preserves the structural component of the source representation while transplanting the concept-specific deviations of the target. The synthetic representation is injected into the classifier, and faithfulness is quantified by measuring the agreement between the predictions of the synthetic example and those of the target.

In a complementary evaluation, we assess whether concept-level interventions can correct misclassified samples. For each misclassified instance, we apply targeted interventions to the concept coefficients by adding or removing specific concepts to support the flipping operation. The goal is to flip the prediction to the correct label and report the resulting correction accuracy.

The results of both evaluations are presented in Table \ref{tab:faithfulness}. The concept-transfer experiment yielded 87.34\% accuracy while the intervention on misclassified samples achieved a 99.8\% success rate. These results suggest that the concept coefficients successfully capture the specific features the model uses for classification. Specifically, cloning these coefficients onto a different base representation reliably results in the target prediction. Furthermore, by performing targeted interventions on these coefficients, one can successfully steer the model’s internal logic toward correcting misclassified samples. These two observations provide strong empirical evidence that the identified concepts are the primary drivers of the final output.

\begin{table}[ht]
    \centering
    \caption{Concept Faithfulness Evaluation: Accuracy in Concept-Transfer and Correcting Miss-classified samples.}
    \label{tab:faithfulness}
    \resizebox{0.97\linewidth}{!}{
    \begin{tabular}{ccc}
        \toprule
        & Concept-Transfer & Correcting Misclassified \\
        \midrule
        Accuracy: & 87.34\% & 99.8\% \\
        \bottomrule
    \end{tabular}
    }
\end{table}

\section{Local Concept-Based Explanations}
\label{sec:local-explanations}
\begin{figure}
    \centering
    \includegraphics[width=\linewidth]{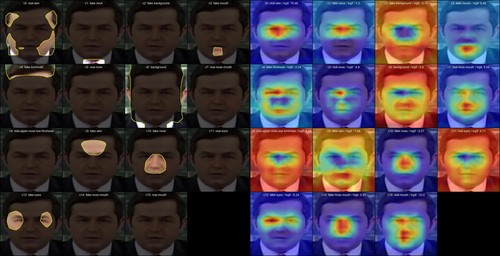}
    \caption{
    Visualization of Concept Presence Maps (left) and the corresponding Concept Contribution Maps (right). 
    Presence maps indicate the spatial activation of each learned concept, while contribution maps highlight their influence on the model's prediction.
    }
    \label{fig:cpm-ccm}
\end{figure}

\begin{figure}
    \centering
    \includegraphics[width=0.9\linewidth]{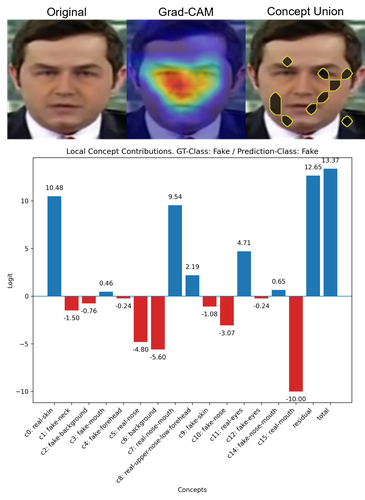}
    \caption{Visualization of the original image, it's Grad-CAM heatmap and the union of all Concept Presence Maps (top). Logit decomposition into individual concept contributions and the residual term (bottom)}
    \label{fig:cam-bar}
\end{figure}

To understand how individual concepts influence a specific classification, we decompose the model's logit with the help of CCMs. We focus on a test sample with a ground truth label of "Fake", which the model correctly classifies.

We employ CCMs to visualize the spatial influence of each concept on the final decision. In Figure \ref{fig:cpm-ccm} , we depict the concept presence maps (CPMs) (left) with their corresponding CCMs (right). We observe that the heatmaps are highly localized, for instance, the activations of $c_{12}$ and $c_3$ are highly activated for regions around the eyes and mouth respectively. This confirms that the model is reacting to specific artifacts within these facial components rather than some global noise. 

Furthermore in Figure \ref{fig:cam-bar} we provide the Grad-CAM \cite{selvaraju2017grad} heatmap, the CPMs union, and the quantitative decomposition of the prediction where we decompose the total logit into individual concepts contributions and a residual term that is equivalent to the proportion of logits not explained by any concept.

\section{What-if-Analysis: Counter-Factual Explanations}
\label{sec:counterfactual}

\begin{figure}
    \centering
    \includegraphics[width=0.9\linewidth]{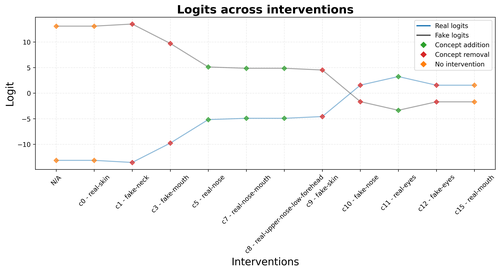}
    \caption{Logit shift under counterfactual concept intervention. We illustrate the causal impact of specific concepts on a single test sample. By starting from a baseline (N/A) and sequentially adding (green) or removing (red) semantic concepts $c_i$, we observe the dynamic shift in model logits}
    \label{fig:counterfactual_example}
\end{figure}

\begin{figure}
    \centering
    \begin{subfigure}[t]{\linewidth}
        \centering
        \includegraphics[width=\linewidth]{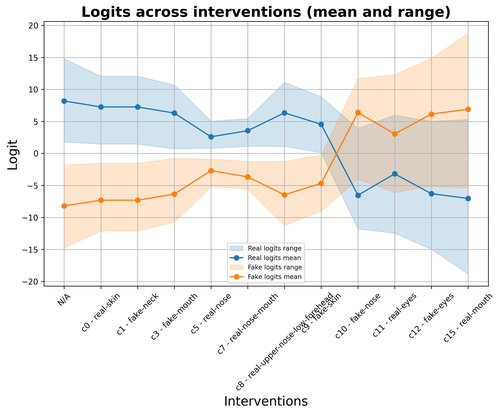}
    \end{subfigure}
    \begin{subfigure}[t]{\linewidth}
        \centering
        \includegraphics[width=\linewidth]{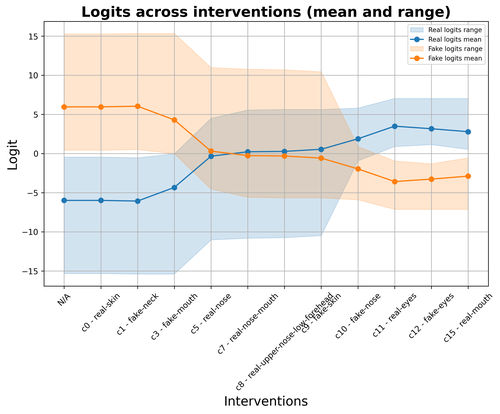}
    \end{subfigure}
    \caption{Statistical distribution of logits across sequential interventions. We report the mean (solid lines) and range (shaded areas) of logits for subsets of samples initially predicted as "Real" (top) and "Fake" (bottom)}
    \label{fig:counterfactual_intervention_stats}
\end{figure}

To evaluate the causal influence of our identified concepts on the model's final decision, we perform a series of counterfactual interventions. By systematically adding or removing concepts, we can observe whether the model's prediction "flips".

\subsection{Intervention methodology}

We define a set of candidate concepts for intervention based on their RCAV sensitivity scores. Specifically, any concept $c_i$ whose absolute sensitivity score satisfies Eq.~\ref{eq:counterfactual-rcav-sens} is considered a primary driver of the model's prediction and thus a candidate for manipulation. Formally,
\begin{equation}
\label{eq:counterfactual-rcav-sens}
|S(c_i)| > \tau,
\end{equation}
where $\tau = 0.9$ and $S(c_i) \in \mathbb{R}$ denotes the scalar RCAV sensitivity score associated with concept $c_i$.

Prior to intervention, we compute concept-specific activation statistics. 
For each concept, we estimate representative signal values using distributional statistics and compute the corresponding top and bottom quantiles, which are later used during intervention, similar to~\cite{doumanoglou2025learning}.

Interventions are performed by modifying the latent representation along the direction of the selected candidate concepts. We consider three types of interventions:

(i) \textbf{Concept addition}, where we randomly sample a mask from samples in which the concept is active and substitute the concept-related information using the top-quantile statistics to strengthen its presence; 
(ii) \textbf{Concept removal}, where we suppress a concept associated with the opposing class by substituting the representation with the bottom-quantile statistics; and 
(iii) \textbf{No intervention}, where the latent representation remains unchanged if a high-sensitivity concept is already present (or absent) in a manner consistent with the target class.

\subsection{Single-Sample Intervention Trace}

Figure~\ref{fig:counterfactual_example} illustrates a successful counterfactual trajectory for a sample initially classified as ``Fake''. We apply a sequence of interventions, beginning with the removal of fake-associated concepts and followed by the addition of real-associated concepts.

As shown in the logit plot, the initial state exhibits a high ``Fake'' logit and a low ``Real'' logit. By removing concepts such as $c_3$ (fake-mouth) and $c_9$ (fake-skin), and adding $c_{11}$ (real-eyes), each intervention progressively shifts the logits toward the target direction, ultimately flipping the prediction. This step-wise transition demonstrates that the model’s decision is strongly tied to the identified concepts.

\subsection{Statistical Robustness of Interventions}

To verify that the observed counterfactual behavior is not limited to a single example, we repeat the analysis on a balanced set of 20 samples: 5 real (ground-truth real, predicted real), 5 real (ground-truth real, predicted fake), 5 fake (ground-truth fake, predicted fake), and 5 fake (ground-truth fake, predicted real). Figure~\ref{fig:counterfactual_intervention_stats} reports the mean and range of logit changes across all samples, grouped by their initial prediction.

For samples initially classified as ``Real'', interventions consistently decrease the mean ``Real'' logit while increasing the ``Fake'' logit. Conversely, samples initially classified as ``Fake'' exhibit the opposite trend, being steered toward ``Real''. The consistency of these trends, reflected in the shaded ranges, suggests that the identified concepts are robust across samples and that the model systematically relies on them to distinguish between real and manipulated faces.

\section{Discussion and Conclusion}
\label{sec:conclusion}

In this study, we applied the EDDP method to provide post-hoc interpretability for deepfake detectors. We successfully identified a semantic vocabulary of internal concepts, effectively mapping abstract latent activations to interpretable human-understandable semantics.

Our faithfulness assessments confirmed that the learned concept directions unveil the internals of the model's decision-making process. The ability to reliably reproduce target predictions by changing their coefficients suggests they capture the essential characteristics used for classification. This is further supported by the CCMs, where concepts and their logit contributions are grounded to facial areas. Finally, the counterfactual "what-if" analysis provides causal evidence for the role of these concepts. By systematically manipulating concept presence and observing the resulting shifts in prediction logits, we demonstrated that the model’s internal logic can be effectively steered. This series of experiments add a level of transparency that is crucial for deepfake detectors.

Despite these strengths, several limitations define the current scope of this work. First, while EDDP does not require retraining the underlying deepfake detector, learning of the EDDP is still required on the representation space of the model to unveil the encoding and decoding directions of concepts. Second, the number of concepts remains a manually defined hyperparameter which may lead to redundant or insufficient signal directions if not selected correctly. Lastly, the identified concepts are inherently optimized for the specific model architecture and dataset used during training. These semantic directions are likely not directly transferable across different models or datasets, meaning a separate training process may be required for each model-dataset combination.

This work demonstrates that the "black-box" nature of deepfake detection can be unraveled through concept-based interpretability. By grounding classification logic in identifiable facial regions and providing a mechanism to steer model predictions we enable more transparent and trustworthy detection.

\section{Acknowledgments}

This research has been supported by the European Commission funded program DETECTOR, under Horizon Europe Grant Agreement 101225942.

{
    \small
    \bibliographystyle{ieeenat_fullname}
    \bibliography{main}
}

\end{document}